\pdfoutput=1
\documentclass[11pt]{article}
\usepackage{EMNLP2023}
\usepackage{times}
\usepackage{latexsym}
\usepackage[T1]{fontenc}
\usepackage[utf8]{inputenc}
\usepackage{microtype}
\usepackage{inconsolata}
\usepackage{float}
\usepackage{graphicx}
\usepackage{multirow}
\usepackage{xcolor}
\usepackage{amsmath}
\usepackage{amsfonts}
\usepackage{booktabs}
\usepackage{placeins}
\usepackage{enumitem}
\usepackage[hang,flushmargin]{footmisc}
\usepackage{amssymb}
\usepackage{pifont}

\newcommand{\cmark}{\ding{51}}
\newcommand{\xmark}{\ding{55}}
\definecolor{er}{HTML}{ff0000}

\title{IdEALS: Idiomatic Expressions for Advancement of Language Skills}

\author{Narutatsu Ri$^1$, Bill Sun$^1$, Sam Davidson$^2$, Zhou Yu$^1$ \\
        $^1$Columbia University \quad $^2$University of California, Davis \\
        \texttt{\{wl2787, bys2017, zy2461\}@columbia.edu} \\
        \texttt{ssdavidson@ucdavis.edu}}

\begin{document}
\maketitle

\begin{abstract}
Although significant progress has been made in developing methods for Grammatical Error Correction (GEC), addressing word choice improvements has been notably lacking and enhancing sentence expressivity by replacing phrases with advanced expressions is an understudied aspect. In this paper, we focus on this area and present our investigation into the task of incorporating the usage of idiomatic expressions in student writing. To facilitate our study, we curate extensive training sets and expert-annotated testing sets using real-world data and evaluate various approaches and compare their performance against human experts.
\end{abstract}

\section{Introduction}
Grammatical Error Correction is a vital task that looks to improve student writing skills by addressing syntactic errors in written text, and various methods have been developed to correct syntactic errors (\citealp{10.5555/3504035.3504741, omelianchuk-etal-2020-gector, https://doi.org/10.48550/arxiv.2104.02310}, among others). However, there remains a gap in the field for improving writing skills through \emph{enhancing} the naturalness and quality of learner writing by replacing grammatically correct phrases with semantically equivalent but advanced expressions. Hence, existing tools have been found inadequate in providing such suggestions \cite{article_grammarly}.

To address this gap, we consider the task of providing replacement suggestions that incorporate idiomatic expressions (IEs) (Figure \ref{fig:sample}). Prior research proposes idiomatic sentence generation (ISG) as the task of transforming sentences into alternative versions including idioms \citep{zhou-etal-2021-pie}, and preserving semantic content has been extensively studied in paraphrase generation (\citealp{mckeown-1979-paraphrasing, wubben-etal-2010-paraphrase, prakash-etal-2016-neural}, etc.). However, such methods are not designed specifically for writing improvement, and standard sequence-to-sequence approaches have been proven ineffective for the ISG task. Additionally, current evaluation metrics are not directly applicable to writing improvement, emphasizing the necessity for methods tailored for this purpose.

In this work, we introduce \emph{Idiomatic Sentence Generation for Writing Improvement} (ISG-WI) as a distinct task for writing improvement under the ISG task. We enhance existing datasets for ISG specifically for word choice recommendation by creating a training set with a broader range of idiomatic expressions and associated training data along with a testing set comprising real-world student-written sentences annotated by human experts, which we refer to as the \emph{Idiomatic Expressions for Advancement of Language Skills} (IdEALS) dataset. We provide precise definitions for performance metrics and explore two different approaches to the ISG-WI task.\footnote{Dataset and code can be found at \url{https://github.com/narutatsuri/isg_wi}.} 

We summarize the main contributions of this work as follows:

\begin{figure}[t]
\centering
\includegraphics[width=\columnwidth]{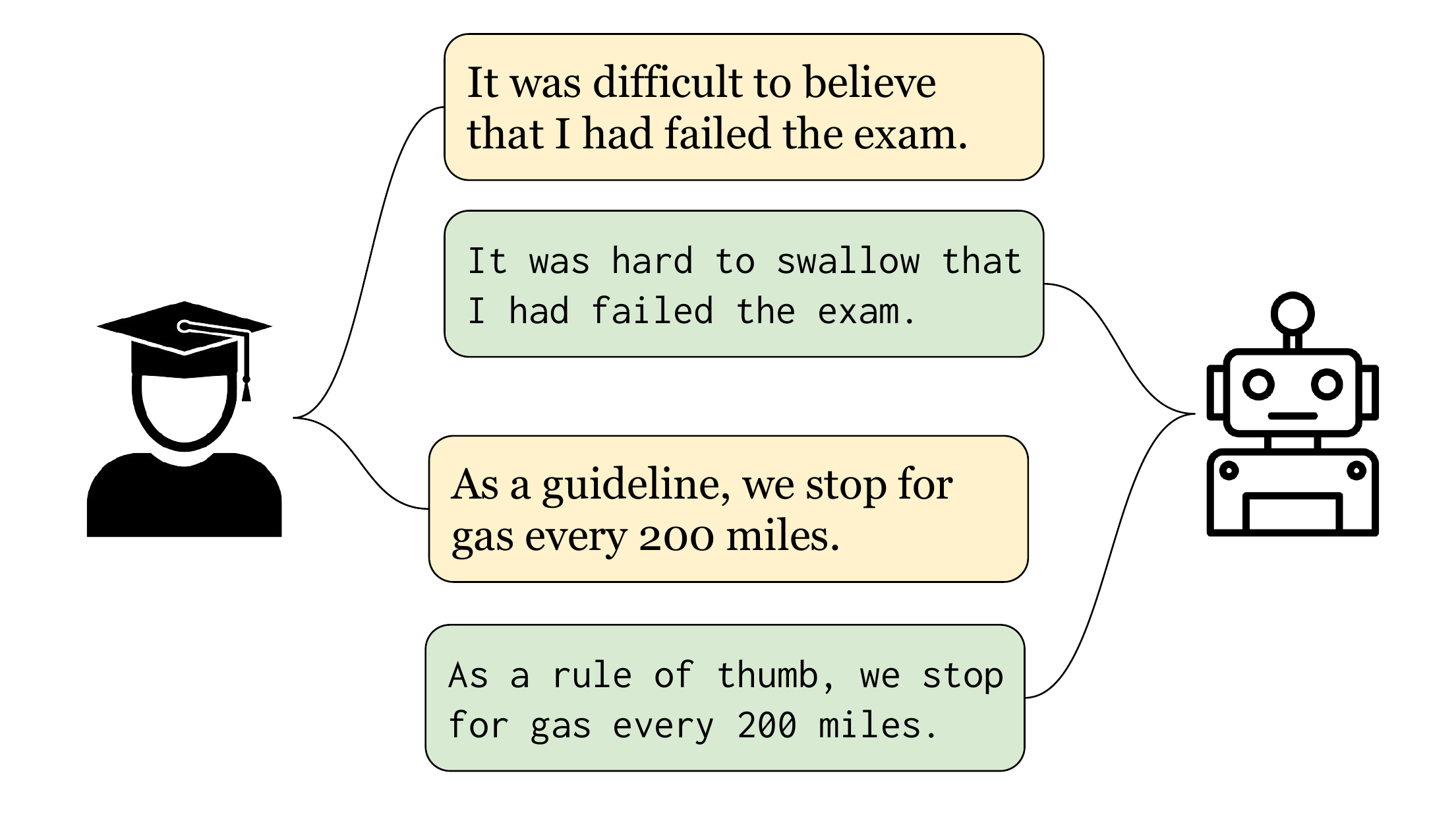}
\small{\caption{\label{fig:sample}Examples of the ISG-WI task. Given a student-written sentence, the task seeks to enhance word choice usage by providing replacement suggestions for words or sentence constituents with idiomatic expressions. }}
\end{figure}
\begin{enumerate}[noitemsep,topsep=0pt]
\item We compile a dataset consisting of a large-scale training set and an expert-annotated test set comprised of real student-written essays.
\item We propose precise metrics to evaluate the performance of the ISG-WI task, and benchmark two approaches against expert human annotators.
\end{enumerate}

\section{Related Work}\label{sec:related_work}
\textbf{Grammatical Error Correction}. Recent studies in Grammatical Error Correction (GEC) have focused on two main directions: generating synthetic data for pre-training Transformer-based models \citep{kiyono-etal-2019-empirical, grundkiewicz-etal-2019-neural, zhou-etal-2020-improving-grammatical} and developing models specifically designed for correcting grammar errors \citep{chollampatt-etal-2016-adapting, nadejde-tetreault-2019-personalizing}. In a related line of research, \citet{zomer-frankenberg-garcia-2021-beyond-grammatical} explore writing improvement models that address errors resulting from the student's native language, going beyond surface-level grammatical corrections. Our work shares a similar objective in aiming to move beyond syntactic mistakes and instead promote advanced word choice usage.

\textbf{Paraphrase Generation}. Paraphrase models often draw inspiration from machine translation techniques \citep{wubben-etal-2012-sentence, mallinson-etal-2017-paraphrasing, https://doi.org/10.48550/arxiv.1709.05074, prakash-etal-2016-neural, https://doi.org/10.48550/arxiv.2006.05477, https://doi.org/10.48550/arxiv.2010.12885} with some efforts devoted to domain-specific paraphrasing \citep{https://doi.org/10.48550/arxiv.1711.00279, https://doi.org/10.48550/arxiv.1808.04364}. Various paraphrase datasets have been curated for training sequence-to-sequence paraphrase models \citep{dolan-brockett-2005-automatically, ganitkevitch-etal-2013-ppdb, https://doi.org/10.48550/arxiv.1702.03814}. Notably, \citet{zhou-etal-2021-pie} introduce the ISG task and propose a corresponding dataset. Building upon their work, we focus specifically on ISG for writing improvement by constructing a larger training dataset and an expert-annotated testing dataset using real-world student-written text and investigate approaches for the ISG-WI task.

\section{The IdEALS Dataset}\label{sec:data_collection}
Here, we detail our procedure for curating the training and testing set. A comparison with the PIE dataset \cite{zhou-etal-2021-pie} is provided in Table \ref{table:dataset_comparison}. 

\begin{table}
\centering
\scalebox{0.8}{
\begin{tabular}[t]{l|cc}
\hline \textbf{Phrase Type} & \textbf{Phrases} & \textbf{Size} \\ \hline
Idioms (Mono.) & 184 & 1722 \\
Idioms (Poly.) & 90 & 1005 \\
Phrasal Verbs & 189 & 1944 \\
Prep. Phrases & 505 & 4156 \\
\hline
Total & 968 &8827 \\
\hline
\end{tabular}}
\scalebox{0.8}{
\begin{tabular}[t]{c|l}
\hline 
\textbf{Changes}  & \textbf{Count} \\ 
\hline
$0$ & 454 \\
$1$ & 326 \\
$2$ & 28 \\
$3$ & 2 \\
\hline
Total & 810 \\
\hline
\end{tabular}}
\small{\caption{\label{table:data_collection}Corpus statistics for each dataset in the PIES corpus. Left and right tables correspond to the training and testing set respectively. "Idioms" refer to idiomatic expressions, and "Phrases" denote the number of phrases each type contains.}}
\end{table}

\begin{table}
\centering
\scalebox{0.9}{
\begin{tabular}[t]{c|cc|cc}
\hline
\textbf{Dataset} & \textbf{Train} & \textbf{Phrases} & \textbf{Test} & \textbf{Real Data} \\
\hline
PIE & 3,524 & 823 & 1,646 & \xmark \\
IdEALS & 8,827 & 968 & 810 & \cmark \\
\hline
\end{tabular}}
\small{\caption{\label{table:dataset_comparison} Comparision of the PIES dataset \cite{zhou-etal-2021-pie} and the IdEALS dataset. "Train," "Test" respectively refer to train and test data size. "Real Data" indicates whether the test data utilizes real-world student-written sentences.}}
\end{table}

\subsection{Training Set}
The training set of the IdEALS dataset comprises sentence pairs consisting of an original sentence and a corresponding paraphrased sentence, where a subpart of the original is replaced with an idiomatic expression where appropriate. Dataset statistics can be found in Table \ref{table:data_collection}.

\textbf{IE Collection}. The training set includes a collection of idioms, prepositional phrases, and phrasal verbs as potential replacements. To curate idioms, we leverage the EPIE dataset \citep{saxena2020epie}, which contains 358 static idiomatic expressions. After removing expressions unsuitable for written essays (e.g., "you bet"), we extract 274 idioms and gather synonyms for each idiomatic expression by scraping online sources. As no publicly available datasets exist for phrasal verbs and prepositional phrases, we obtain 1000 phrasal verbs and 633 prepositional phrases from online educational sources\footnote{\raggedright Phrasal verbs: \url{https://www.englishclub.com/store/product/1000-phrasal-verbs-in-context/}\\ Prepositional phrases: \url{https://7esl.com/prepositional-phrase/}}. We then filter out trivial phrases (e.g., "calm down") to extract 189 phrasal verbs and 505 prepositional phrases, each accompanied by example sentence usages.

\textbf{Sentence Pair Generation}. For idioms, we utilize the example sentences provided in the EPIE dataset. We create parallel sentence pairs by replacing idioms in the examples with synonyms, which serve as the original sentences. As the number of example sentences for phrasal verbs and prepositional phrases is limited compared to idioms, we address this by adopting in-context learning methods \citep{https://doi.org/10.48550/arxiv.2005.14165} and construct prompts for large language models to generate additional example sentences. Subsequently, we manually verify each sentence pair to ensure grammatical correctness and semantic consistency between the original and paraphrased pairs. Prompt details are provided in Appendix \ref{app:setence_generation}. 

\subsection{Testing Set}
In order to assess the effectiveness of ISG-WI methods on student-written text, it is essential to utilize real-world student-written sentences that may contain grammatical issues as testing data to ensure that the testing set accurately reflects the challenges and characteristics of student writing. In light of this, we construct the testing set by collecting real sentences written by students and annotating them using human experts.

\textbf{Original Sentences}. We gather 810 sentences from the ETS Corpus of Non-Native Written English \citep{ets}, a curated database of essays written by students for the TOEFL exam. To ensure high-quality annotations, we enlist five graduate students pursuing linguistics degrees as annotators for this task.

\textbf{Annotation Scheme}. The annotators are instructed to preserve both the sentence structure and semantics while providing only idiomatic expressions as suggestions. For each sentence, annotators have the option to provide an alternate sentence in which specific subparts are replaced with an idiomatic expression. If a sentence is deemed not amenable to enhancement, annotators can choose not to provide any annotation. Given that multiple IEs can be equivalent replacements for the same phrase, and some sentences may contain multiple replaceable subparts, we encourage annotators to provide multiple annotations for the same sentence whenever possible. Annotation statistics are included in Table \ref{table:data_collection}. 

\section{Methods}\label{sec:ideals_models}
Previous studies highlight the limitations of vanilla sequence-to-sequence models in preserving semantic content and grammatical fluency for the ISG task \cite{zhou-etal-2021-pie}. In this paper, we explore two methods for the ISG-WI task and assess their performances on the IdEALS testing set.

\textbf{Fine-tuning}. We investigate the use of modern text-to-text models that are fine-tuned on the IdEALS dataset. Specifically, we employ the Parrot paraphraser model \cite{prithivida2021parrot}, which is based on the \texttt{t5-base} model from the T5 model family \cite{https://doi.org/10.48550/arxiv.1910.10683}. By fine-tuning the Parrot model with our training set, we enable the model to generate idiomatic sentence suggestions.

\begin{figure}[t]
\centering
\includegraphics[width=\columnwidth]{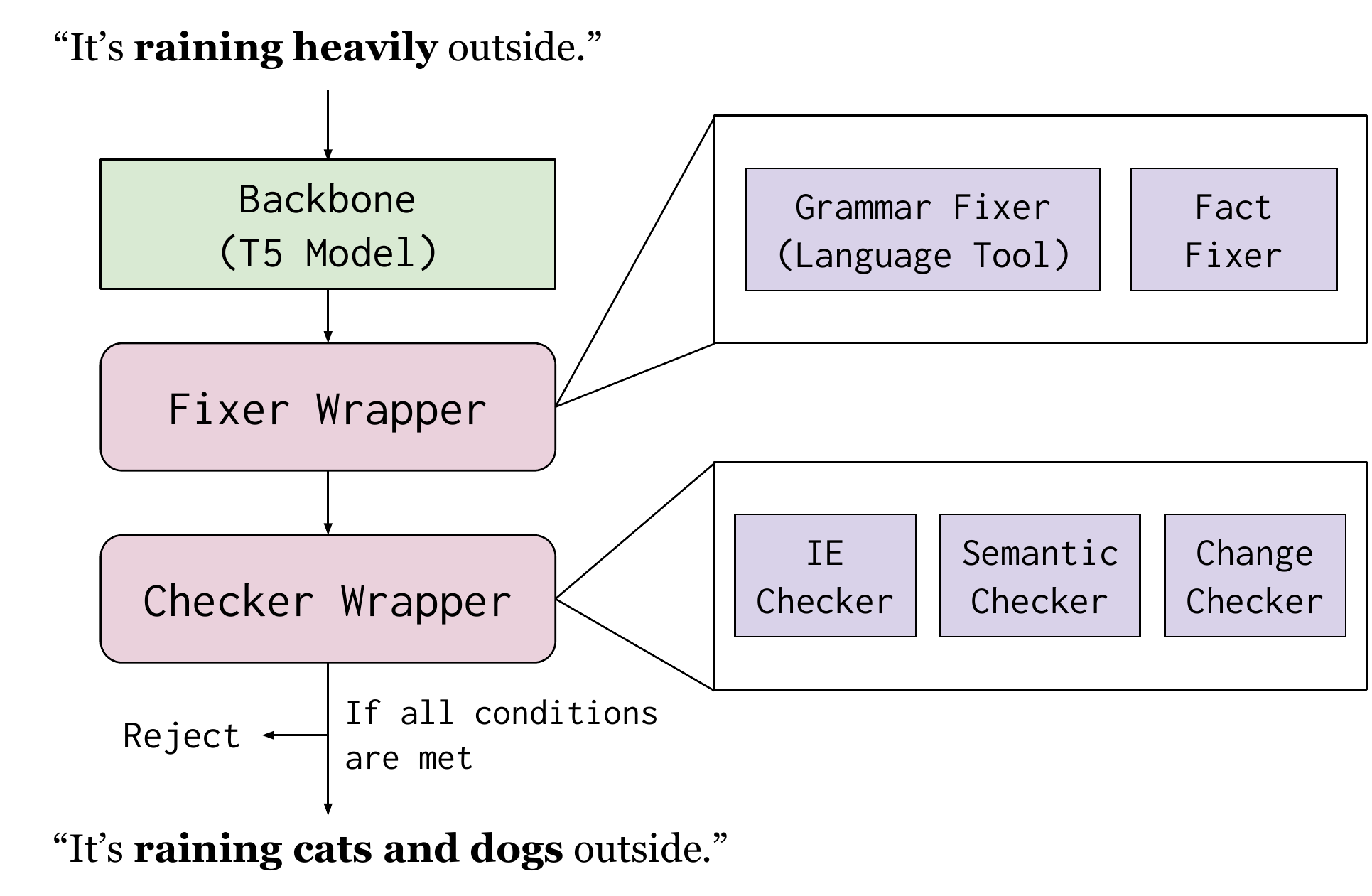}
\small{\caption{\label{fig:model_architecture} Model architecture for the fine-tuned model. Green blocks indicate the usage of neural language models, while purple blocks indicate rule-based heuristics.}}
\end{figure}

To ensure the fulfillment of IdEALS task objectives, we employ postprocessing layers for grammar correction and verification of proper noun hallucinations and incorrect replacements and evaluate whether the task objectives are met with the following criteria. First, we assess semantic consistency between input and output sentences using large language models trained on NLI datasets, where a paraphrase is considered semantically consistent if it exhibits stronger entailment than neutrality. Second, we verify the presence of idiomatic expressions in the output by compiling a comprehensive pool of idioms from online sources. Finally, we assess the preservation of sentence structure by confirming that only one subpart of the original sentence is replaced in the paraphrase. Figure \ref{fig:model_architecture} presents a diagram illustrating the architecture of our proposed models. Additional model specifics and training details are provided in Appendix \ref{app:model_details}.

To assess the efficacy of both the existing datasets and our model architecture, we train the backbone model using the PIE dataset \cite{zhou-etal-2021-pie} and the IdEALS dataset and compare the performances of these trained models.

\begin{table*}[t]
\centering
\begin{tabular}{l|c|c|cccc}
\hline
\multirow{2}{*}{\textbf{Annotator}} & \multirow{2}{*}{\textbf{No Edits}} & \multirow{2}{*}{\textbf{Good Edits}} & \multicolumn{4}{c}{\textbf{Bad Edits}} \\
 & & & \textbf{Adeq.} & \textbf{Corr.} & \textbf{Rich.} & \textbf{Total}  \\
\hline
Parrot + PIE & 475 & 141 & 83 & 15 & 102 & 194 \\
Parrot + PIE + Wrapper & 671 & 124 & 10 & 0 & 5 & 15   \\
Parrot + IdEALS & 347 & 189 & 138 & 11 & 186 & 274 \\
Parrot + IdEALS + Wrapper & 601 & 182 & 22 & 3 & 4 & 27  \\
\hline
\texttt{gpt-3.5-turbo} & 91 & 461 & 15 & 111 & 191 & 258 \\
\texttt{text-davinci-003} & 394 & 160 & 16 & 4 & 240 & 254 \\
\hline
Human experts & 453 & 311 & 4 & 0 & 32 & 36 \\
\hline
\end{tabular}
\caption{\label{table:performances} Performance of Different Models and Human Experts. "+" denotes dataset usage or the checkbox module. Few-shot in-context learning is conducted with 10 in-context examples.}
\end{table*}

\textbf{In-Context Learning}. We also explore the application of in-context learning as an alternative approach. In-context learning involves keeping the parameters of a pre-trained language model fixed while providing it with a prompt containing task descriptions, examples, and a test instance. By leveraging the context, the model can learn patterns and make predictions based on the given task constraints. We construct the prompt by including task instructions and rules for the model to follow, followed by in-context examples sampled from the training set. 

\section{Results}\label{sec:results}
\subsection{Evaluation Metrics}
In this section, we outline the scoring criteria used to evaluate the quality of annotations, whether generated by models or human annotators, using the testing set. Utilizing existing paraphrase metrics \cite{liu-etal-2010-pem, patil2022understanding, shen-etal-2022-evaluation}, we establish three criteria to assess the quality of an annotation:
\begin{itemize}[noitemsep,topsep=0pt]
\item \textbf{Adequacy}: whether the semantic content of the original sentence is preserved in the output.
\item \textbf{Correctness}: whether the sentence structure is maintained, and no new grammatical mistakes are introduced in the output.
\item \textbf{Richness}: whether the provided suggestion incorporates an idiomatic expression.
\end{itemize}

Annotations that meet all three criteria are considered as \emph{good} annotations, while annotations that violate more than one criterion are classified as \emph{bad} annotations.

\subsection{Performances}
Here, we evaluate the abovementioned methods against annotations from human experts and present the results in Table \ref{table:performances}.

\textbf{Fine-tuning}. 
Comparing the backbone model trained on the PIE dataset and the IdEALS dataset, we observe a significant increase in the number of annotations with the IdEALS dataset. In both cases, there is a reduction in erroneous annotations when postprocessing wrappers are used, but occasionally at the expense of a slight decrease in the number of good annotations.

\textbf{In-context Learning}. The two models exhibit distinct performance results. Notably, \texttt{gpt-3.5-turbo} surpasses human experts in generating good annotations but struggles with maintaining correctness. Conversely, \texttt{text-davinci-003} falls short in producing good annotations compared to fine-tuned models on the IdEALS dataset and frequently violates the richness criteria.

\textbf{Human Experts}. 
We observe that experts exhibit a slight tendency to make richness mistakes but consistently deliver annotations of the highest quality compared to automated methods. 

\subsection{Error Analysis}
Here, we analyze the prominent error cases for both methods. Error examples are included in Table \ref{table:error_case} in the Appendix.

\textbf{Fine-Tuning}. The fine-tuned model exhibits two notable error cases: violating the adequacy condition due to semantic changes and suggesting trivial changes that lack idiomatic expressions. These errors are often observed in input sentences with syntactic errors or inappropriate phrase usage, highlighting the model's vulnerability to mistakes when the input contains errors.

\textbf{In-Context Learning}. In-context learning methods tend to over-replace words with semantically equivalent phrases, resulting in insignificant changes compared to fine-tuned models. Addressing this issue proves challenging, even with additional rules in the prompt.

\section{Conclusion}
We introduced the ISG-WI task as a crucial step towards improving word choice in writing. We curate an extensive training set and real-world student-written texts annotated by human experts for our testing set. Our experimental results demonstrate the effectiveness of our dataset, with language models trained on the IdEALS dataset exhibiting superior performance in generating suggestions for idiomatic expressions compared to existing ISG datasets.

\section*{Ethical Considerations}
Obtaining high-quality annotations for idiomatic sentences poses challenges as it necessitates expertise in language teaching and significant time commitment. We took measures to ensure that annotators were adequately compensated for their valuable contributions.

\bibliography{emnlp2023}
\bibliographystyle{acl_natbib}

\begin{table*}[!htbp]
\centering
\setlength{\tabcolsep}{6pt}
\renewcommand{\arraystretch}{1.2}
\begin{tabular}{p{0.08\textwidth}|p{0.2\textwidth}p{0.7\textwidth}}
\hline
\textbf{Set} & \textbf{Phrase Type} & \textbf{Example Sentence} \\ \hline
\multirow{ 8}{*}{\textbf{Train}} & \multirow{ 2}{*}{Idioms (Mono.)} & $s $:  
\hspace{1mm}We’re not \colorbox{pink}{in sync}. Listen carefully to what I am telling you.	 \\
&& $s'$: We’re not \colorbox{green}{on the same page}. Listen carefully to what I am telling you.  \\
&\multirow{ 2}{*}{Idioms (Poly.)} & $s$: \hspace{1mm}I was \colorbox{pink}{excited} when I found out that I'd gotten a good grade.\\
&& $s'$: I was \colorbox{green}{as high as a kite} when I found out that I'd gotten a good grade.  \\
&\multirow{ 2}{*}{Phrasal Verbs} & $s$: \hspace{1mm}I was \colorbox{pink}{nearly} failing the test when I got the extra credit question.\\
&& $s'$: I was \colorbox{green}{within an ace of} failing the test when I got the extra credit question.\\
&\multirow{ 2}{*}{Prep. Phrases} & $s$: \hspace{1mm}I'm going to \colorbox{pink}{renew} my home before my in-laws come to visit.\\
&& $s'$: I'm going to \colorbox{green}{spruce up} my home before my in-laws come to visit. \\
\hline
\multirow{6}{*}{\textbf{Test}} & & $s$: We become \colorbox{pink}{lethargic} studying something for a long time. \\
 & & $s'$: We become \colorbox{green}{heavy-eyed} studying something for a long time.\\
 & & $s$: So, people must \colorbox{pink}{rely on each other} and must be active.\\
 & & $s'$: So, people must \colorbox{green}{have each others' backs} and must be active.\\
 & & $s$: To sum up, \colorbox{pink}{enjoying life} is not related to the person's age.\\
 & & $s'$: To sum up, \colorbox{green}{making the most out of life} is not related to the person's age.\\
\end{tabular}
{\small\caption{\label{table:example_ideals} Sentence pairs from the training and testing sets. $s$ denotes the original sentence and $s'$ denotes the paraphrased sentence. The training set includes idiomatic expressions: idioms (monosemous, polysemous), phrasal verbs, and prepositional phrases. The original subphrase to be replaced is highlighted in red, and the replacement idiomatic expression is highlighted in green.}}
\vspace*{25pt}
\setlength{\tabcolsep}{6pt}
\renewcommand{\arraystretch}{1.5}
\begin{tabular}{p{0.08\textwidth}|p{0.2\textwidth}p{0.7\textwidth}}
\hline
\textbf{Method} & \textbf{Error Type} & \textbf{Example Sentence} \\ \hline
\multirow{8}{*}{\textbf{FT}}&\multirow{4}{*}{Incorrect Semantics} & $s$: \hspace{1mm} To process a car is like \colorbox{pink}{a liberty} of moving, you can do whatever you want in terms of transportation.\\
&& $s'$: To process a car is like \colorbox{yellow}{a piece of cake} of moving, you can do whatever you want in terms of transportation. \\
&\multirow{4}{*}{Trivial Change} & $s$: They know that if they give a false picture of their product in an advertisement, they might fool customers for a short \colorbox{pink}{term}.\\
& & $s'$: They know that if they give a false picture of their product in an advertisement, they might fool customers for a short \colorbox{yellow}{time}.\\
\hline
\multirow{4}{*}{\textbf{ICL}}& \multirow{4}{*}{Trivial Change} & $s$: By this, I mean that we are all \colorbox{pink}{concentrated} on our projects and we work so hard that we do not have time to think about ourselves.\\
& & $s'$: By this, I mean that we are all \colorbox{yellow}{focused} on our projects and we work so hard that we do not have time to think about ourselves.\\
\hline
\end{tabular}
\caption{\label{table:error_case}Example error cases of fine-tuning and in-context learning methods. "FT" refers to fine-tuned models and "ICL" refers to in-context learning methods. The original replaced subphrase is highlighted in red, and the erroneous replacemet is highlighted in yellow.}
\end{table*}

\appendix
\section{Supplementary Material}
\subsection{Model Details}\label{app:model_details}
The backbone model is trained for 20 epochs using the PIE dataset and the IdEALS dataset on two NVIDIA GeForce RTX 3090 graphics cards. For the postprocessing wrappers, we utilize the LanguageTool API\footnote{https://languagetool.org/} for the Grammar Fixer module. The Fact Fixer module utilizes the \texttt{bert-base} model trained on the CoNLL-2003 Named Entity Recognition dataset \cite{tjong-kim-sang-de-meulder-2003-introduction}, while the Semantic Checker module employs the \texttt{roberta-large} model trained on the Stanford NLI dataset \cite{bowman-etal-2015-large}, the MultiNLI dataset \cite{williams-etal-2018-broad}, and the Adversarial NLI dataset \cite{nie-etal-2020-adversarial}. Both models are accessible at \url{huggingface.co}.

\subsection{Dataset Samples}\label{app:example_ideals}
Examples from both the training and testing sets can be found in Table \ref{table:example_ideals}. Each idiomatic expression type collected in the training set is represented by one example. Note that the test set does not have the idiomatic expression type labeled. 

\subsection{Prompt for Sentence Pair Geneneration}\label{app:setence_generation}
Here, we include the prompt utilized for generating additional training samples for phrasal verbs and prepositional phrases. \\

{\tiny\ttfamily 
\noindent [TASK DESCRIPTION] Generate 5 example sentences that use the PHRASE provided without altering the semantic content of the original sentence.\\\\
Example \#1: \\
PHRASE: wet behind the ears\\
SENTENCES: \\
- terry, it turned out, was just out of university and wet behind the ears.\\
- the song is all about how he felt as a small town, wet behind the ears kid coming to la for the first time.\\
- hawking was a research student, still wet behind the ears by scientific standards.\\\\
Example \#2: \\
PHRASE: narrow down\\
SENTENCES:\\
- I can't decide what to wear, so I'm going to narrow down my options to three dresses.\\
- We only have a limited amount of time, so we need to narrow down our options.\\
- After doing some research, I was able to narrow down my list of colleges to five.\\\\
Example \#3: \\
PHRASE: bounce back \\
SENTENCES:\\
- After his divorce, he was able to bounce back and start dating again.\\
- The company's sales took a hit after the recession, but they were able to bounce back and return to profitability.\\
- He was disappointed when his team lost the championship game, but he was able to bounce back and win the next one.\\\\
{-}{-}{-}{-}{-}{-}{-}{-}{-}{-}{-}{-}{-}{-}{-}{-}{-}{-}{-}{-}{-}{-}PROMPT ENDS HERE{-}{-}{-}{-}{-}{-}{-}{-}{-}{-}{-}{-}{-}{-}{-}{-}{-}{-}{-}{-}{-}{-}\\
PHRASE: \\
SENTENCES:\\
\par}

\subsection{Prompt for In-Context Learning}\label{app:example_prompts}
The full prompt for few-shot settings is included, consisting of 10 examples sampled from our training set. An example completion by the \texttt{text-davinci-003} model is provided at the end.\\

{\tiny\ttfamily 
\noindent [TASK DESCRIPTION] Enhance the input sentence by identifying phrases that can be replaced with a potentially idiomatic expression and output the replaced input sentence. \\
Rules:\\
1. Do not change the sentence besides replacing one phrase. \\
2. Only replace phrases with idiomatic expressions.\\
3. Do not alter the semantic content of the original sentence.\\
4. If there are no phrases replaceable, return "nan".\\\\
Example \#1: \\
INPUT: The city was besieged for weeks, and the people were running out of food and water.\\
OUTPUT:  the city was under siege for weeks, and the people were running out of food and water.\\\\
Example \#2: 
INPUT:  No matter what, we must ensure that our children have a bright future.\\
OUTPUT:  at all costs, we must ensure that our children have a bright future.\\\\
Example \#3: \\
INPUT:  The train was travelling rapidly 200 kilometers per hour.\\
OUTPUT:  the train was travelling at a speed of 200 kilometers per hour.\\\\
Example \#4: \\
INPUT: Wow, man, this party is great!\\
OUTPUT: wow, man, this party is out of sight!\\\\
Example \#5: \\
INPUT:  In order to achieve success, she was willing to work long hours at the cost of her social life. \\
OUTPUT:  in order to achieve success, she was willing to work long hours at the expense of her social life. \\\\
Example \#6: \\
INPUT: We hope that by forming a bipartisan committee we will be able form a body that represents the most ideal circumstances.\\
OUTPUT: we hope that by forming a bipartisan committee we will be able form a body that represents the best of both worlds.\\\\
Example \#7: \\
INPUT:  The best way to eliminate the undesirables is to set high standards.\\
OUTPUT:  The best way to weed out the undesirables is to set high standards.\\\\
Example \#8: \\
INPUT:  The top prize in the raffle is unclaimed.\\
OUTPUT:  the top prize in the raffle is up for grabs.\\\\
Example \#9: \\
INPUT:  The news of the government's corruption was the last straw, and people finally began to explode in protest.\\
OUTPUT:  The news of the government's corruption was the last straw, and people finally began to break out in protest.\\\\
Example \#10: \\
INPUT:  He refused to surrender even when faced with overwhelming odds.\\
OUTPUT:  He refused to back down even when faced with overwhelming odds.\\\\
{-}{-}{-}{-}{-}{-}{-}{-}{-}{-}{-}{-}{-}{-}{-}{-}{-}{-}{-}{-}{-}{-}PROMPT ENDS HERE{-}{-}{-}{-}{-}{-}{-}{-}{-}{-}{-}{-}{-}{-}{-}{-}{-}{-}{-}{-}{-}{-}\\
INPUT: All people in a society can not be happy with the conditions or lifestyles that they are living in.\\
OUTPUT: All people in a society can not be on cloud nine with the conditions or lifestyles that they are living in. \par}

\end{document}